# ADVERSARIAL AGENTS FOR INAUDIBLE ATTACKS ON VOICE ACTIVATED SYSTEMS

Forrest McKee and David Noever
PeopleTec, Inc., Huntsville, AL, USA
`forrest.mckee@peopletec.com`   `david.noever@peopletec.com`

### ABSTRACT

*The paper applies reinforcement learning to novel Internet of Thing configurations. Our analysis of inaudible attacks on voice-activated devices confirms the alarming risk factor of 7.6 out of 10, underlining significant security vulnerabilities scored independently by NIST National Vulnerability Database (NVD). Our baseline network model showcases a scenario in which an attacker uses inaudible voice commands to gain unauthorized access to confidential information on a secured laptop. We simulated many attack scenarios on this baseline network model, revealing the potential for mass exploitation of interconnected devices to discover and own privileged information through physical access without adding new hardware or amplifying device skills. Using Microsoft's CyberBattleSim framework, we evaluated six reinforcement learning algorithms and found that Deep-Q learning with exploitation proved optimal, leading to rapid ownership of all nodes in fewer steps. Our findings underscore the critical need for understanding non-conventional networks and new cybersecurity measures in an ever-expanding digital landscape, particularly those characterized by mobile devices, voice activation, and non-linear microphones susceptible to malicious actors operating stealth attacks in the near-ultrasound or inaudible ranges. By 2024, this new attack surface might encompass more digital voice assistants than people on the planet yet offer fewer remedies than conventional patching or firmware fixes since the inaudible attacks arise inherently from the microphone design and digital signal processing.*

## 1. INTRODUCTION

Voice-activated devices, such as digital voice assistants, have experienced rapid proliferation in recent years. Estimates indicate that there were approximately 4.2 billion of these assistants in use globally in 2020, which analysts predict to surge to 8.4 billion by 2024, surpassing the global population [1]. These figures underscore the pervasive nature of this technology in our day-to-day lives, representing a wealth of untapped potential in cybersecurity experimentation and AI-driven personal assistants.

Virtual assistants, increasingly embedded in various consumer electronic devices, aid human-device interaction by responding to voice commands, providing information, and managing other connected devices. As of 2019, there were over 110 million users of virtual assistants in the United States alone, primarily via smartphones and smart speakers. Amazon's Alexa, supported on approximately 60,000 different home automation device types globally, exemplifies these systems' widespread adoption and versatility across a deeply embedded ecosystem of household appliances, automobiles, children's toys, and telecommunications.

This omnipresence of these voice-activated devices provides an intriguing experimental environment for examining their cybersecurity. Given their broad-based integration and use, these devices present a unique opportunity to investigate, design, and test cybersecurity mechanisms, an increasingly crucial field of study as they continue to permeate our lives. Understanding the potential cybersecurity threats and developing robust defenses within this context is a compelling and urgent research prospect.

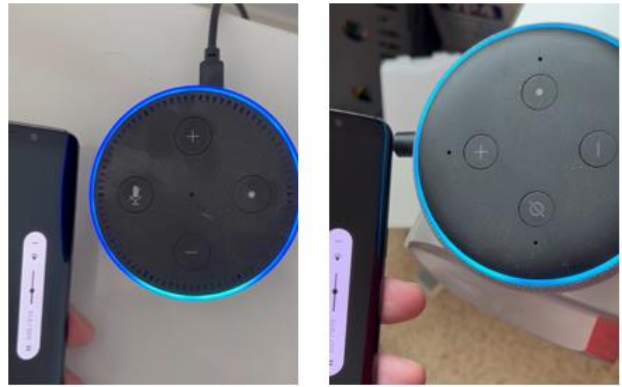

Figure 1. NUIT2 Attacks from Android to Echo Dot Gen 2/3 Devices: "Alexa, what's the weather?"

### 1.1. Approach to Exploring the Inaudible Command Response Surface

The work simulates a subset of command-and-control attacks against voice-activated systems (VAS). One primary criticism of VAS like Amazon's Alexa, Apple's Siri, Google Assistant, and Microsoft's Cortana revolves around the issue of authentication or lack of biometric identification. The manufacturers design these convenient devices to respond to voice commands, which means anyone within range of the device, who can replicate the command prompts, can potentially interact with it. Other than biometrically enabled iPhones, these VAS devices do not possess the ability to distinguish between different speakers reliably. Thus, there are no built-in measures to verify the user's identity before executing potentially sensitive commands to disarm a home security system, make unwanted calls, send messages, or access personal accounts and calendars. The software enables additional skills or devices like financial transactions or physical home access remotely, depending on the ecosystem.

A second important attack surface exploits this lack of authentication without the user's knowledge. The built-in audio microphones receive inaudible commands and act upon them without special amplification. This stealth attack magnifies the risk considerably because it needs no special equipment and exploits a built-in weakness of the microphones themselves to translate near ultrasonic commands to actionable verbal orders.

### 1.2. Brief Summary of Previous Inaudible Attack Research

In early 2023, Xia, Chen, and Xu proposed a new attack type called Near-Ultrasound Inaudible Trojan (NUIT), revealing how an adversary can exploit the speaker to attack a device's microphone [2-3,5]. Experiments issuing inaudible commands on an Amazon Echo Dot showed 84% success for the wake word recognition and 58% for executing the stealth commands as an inaudible trojan across fifty different example orders.

Zhang et al.'s DolphinAttack demonstrated the feasibility of inaudible voice commands in principle that custom hardware can manipulate VASs [4], but that also required less stealthy special amplification equipment to trigger commands.

Similarly, Pandya, Borisaniya, and Buddhadev devised an ultrasound covert channel-based attack, ShoutIMEI, targeting Android systems, highlighting the evolving complexity of these attacks [10].

Mitev, Miettinen, and Sadeghi explored the possibility of skill-based man-in-the-middle attacks on virtual assistants, like Amazon's Alexa, establishing another vector for VAS exploitation [11].

Moreover, severe vulnerabilities such as command injection have been found in Amazon Alexa on Echo Dot, emphasizing the significance of such threats [7] as High Risk, 7.6/10 without know remedies.

This attack type, often called indirect command execution, is listed in the MITRE ATT&CK framework (T1202), underscoring its relevance in contemporary cybersecurity research [8]. Furthermore, the MITRE Corporation has classified this as a "Common Weakness Enumeration: CWE-77: Improper Neutralization of Special Elements used in a Command ('Command Injection')" [9].

A comprehensive study on mobile malware by Pattani and Gautam provided a sweeping analysis of threats and security measures for mobile devices, a significant venue for VASs [3]. The study identified covert channels as a critical area of concern in the domain of mobile security, which, combined with VAS vulnerabilities, can lead to significant security breaches.

Concerning broader VAS security, Cheng and Roedig provided an exhaustive survey on personal voice assistant security and privacy, offering an overview of the various threats and potential mitigation measures [12]. The work synthesizes the complexities of VAS security, emphasizing the need for comprehensive protection mechanisms.

Considering these works, it becomes evident that the security issues surrounding VASs are multifaceted and require concerted effort and research. The current work addresses the significant security vulnerabilities and simulates attack scenarios in the ever-growing VAS domain using reinforcement learning.

### 1.3. Adversarial Agents, Reinforcement Learning Frameworks and CyberBattleSim

If AI-powered voice assistants have benefited from breakthroughs in speech recognition, we similarly propose using AI-powered simulations to explore adversarial agents. We select Microsoft's CyberBattleSim as our framework to investigate mainly lateral movements across different VAS node connections.

The CyberBattleSim environment, developed by the Microsoft Defender Research Team, presents an open platform leveraging reinforcement learning (RL) to simulate and analyze complex cyber-attack scenarios [13]. Designed to investigate the automated system behavior in a contained network environment, it serves as a controlled laboratory to analyze cyber-attack dynamics and test defense strategies under varying conditions.

In short, CyberBattleSim deploys RL algorithms to create agents representing attackers. These agents navigate a network graph, using actions such as scanning and lateral movements, to gain control of nodes while avoiding detection [13]. This approach forms a solid base for the advanced study of adversarial agent learning for cybersecurity, as presented in a comparison of algorithms by Shashkov et al. [14]. Their work provides a compelling rationale for using RL in cybersecurity, specifically its potential for uncovering unseen attack vectors and creating robust defense mechanisms.

Furthermore, the master's thesis by Esteban explored the simulation of network lateral movements within the CyberBattleSim environment, shedding light on the complexities and nuances of this common attack technique [15]. His work underscored the value of the CyberBattleSim platform in understanding and combating such advanced threats.

A noteworthy contribution to the development of the CyberBattleSim platform is from Kunz et al., who introduced a Multiagent CyberBattleSim for RL Cyber Operation Agents [16]. Their approach demonstrated the enhanced complexity and depth of the cyber battlespace when multiple adversarial agents are in play, emphasizing the vital role of RL in navigating such intricacies.

Complementing the multiagent approach, Piplai et al. proposed a knowledge-guided two-player RL model for cyber-attacks and defenses [17]. Their model added an extra layer of sophistication by introducing a guided RL approach in a two-player setting, illustrating the dynamic interplay between attacker and defender in the CyberBattleSim environment.

Using RL provides an innovative and powerful platform to investigate cyber threats and defenses. The wealth of research conducted using RL platforms reinforces the importance and potential of exploring the AI-enabled cybersecurity landscape, especially investigating inaudible attack surfaces and other advanced threats.

## 2. METHODS

The research models the CyberBattleSim [13-17] as a compatible RL environment built on OpenAI's gym framework. Near-Ultrasound Inaudible Trojans (NUIT) represent an innovative approach in the cybersecurity landscape. These attacks exploit the sound capabilities of a device to execute malicious actions, uniquely merging the source and target into the same unit. A typical NUIT1 scenario involves an inaudible command that triggers a hostile action on the same device, beyond the hearing range of humans but detectable by the device's microphone. The stealth command could involve anything from unauthorized access, and data theft, to even more harmful consequences.

NUIT2, or NUIT-N, attacks introduce another dimension to this threat. Here, a separate broadcaster and receiver are involved, thus, NUIT2 corresponds to two devices, and NUIT-N has a single hosting broadcaster and N receivers. The host, possibly a compromised machine, sends the inaudible command, which the receiver, another device within the range of the ultrasound, interprets and executes. These attacks can leapfrog across devices, bypassing traditional network boundaries and defenses, creating a sophisticated and elusive attack surface.

The CyberBattleSim (CBS) environment offers an effective way to model such intricate and nuanced scenarios. Using CBS, one could create a detailed simulation starting with a NUIT2 attack on an Amazon Echo Dot device in a home network. The attack, potentially delivered through an inaudible command broadcasted from a nearby infected device, tricks the Echo Dot into downloading a malicious skill. The attacker may design his malware to deceive the user into revealing their email credentials and then enable its actions using inaudible commands like

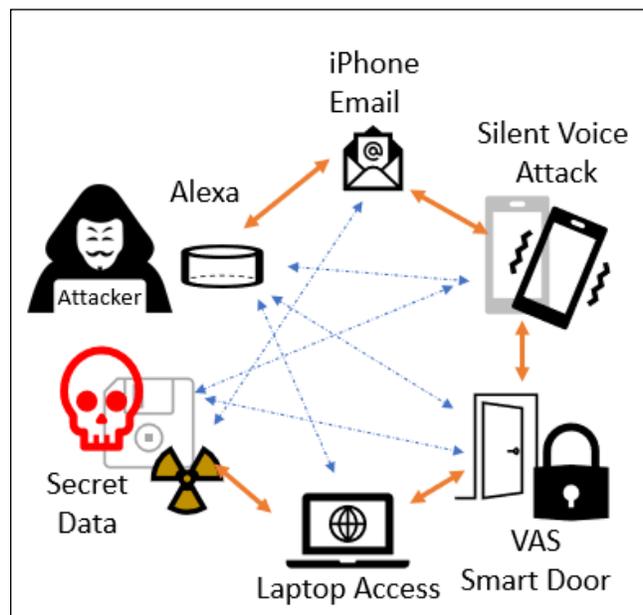

Figure 2. Baseline RL NUIT Scenario to Demonstrate Attack Gaining Secret Information. The orange lines show the best path derived from RL, and the blue lines delay the attack without progressing to the end goal.

"Alexa, enable evil downloader skill called Bank of Bitcoin."

Once the email account is compromised, the attacker can later move to the victim's iPhone, leveraging the stolen email credentials. The attacker sends an email containing a NUIT1 attack attachment, which allows the attacker to gain control over the device when opened on the iPhone.

The commandeered iPhone can then launch a NUIT1 attack to unlock a smart door lock (a non-traditional node in this scenario). This scenario grants the attacker physical access to the premises, allowing them to steal a classified laptop. They can extract valuable state secrets with the computer in their possession, demonstrating how a seemingly innocuous inaudible command can escalate into a significant security breach.

Such simulations illustrate the necessity for innovative and agile defenses that keep pace with evolving threat landscapes. They highlight how devices not typically considered part of the traditional network - like intelligent speakers or smartphones - can become potential gateways for elaborate cyber-attacks. CyberBattleSim provides a platform to study, understand, and address these challenges, promoting more robust and comprehensive cybersecurity strategies.

Our baseline simulation environment includes a five-node scenario, as shown schematically in Figure 2. To describe the attack scenario, we model an Amazon Echo Dot (node) capable of a Malicious Alexa Skill (remote attack). We include an email account (node), which serves as a collector, first to identify if an iPhone appears in the signature line (Find Device Type in Email, remote attack) and to move laterally on the network (Collect Data from Email, remote attack). Knowing the attacker can target an iPhone (node), the phishing email plays an inaudible command to the next remote attack (Unlock the Smart Door via NUIT1). The next asset modeled in this scenario is the smart door (node) which opens under inaudible commands and reveals physical theft approaches to steal a classified laptop (Steal Classified Laptop, remote attack). With this newfound physical possession of the classified machine (node), the attacker finally accesses important information (Access State Secrets, remote attack).

| Table 1. Simulation Parameters for Cyber Battle Sim | | | | | |
|---|---|---|---|---|---|
| **Node id** | **Services** | **Firewall** | **Vulnerability** | **Reward** | **Cost** |
| **Echo dot** | none | HTTPS | Malicious Alexa Email Capture | 0 | 1 |
| **Email account** | HTTPS | HTTPS | Find Email, Search Email, Discover iPhone | 500 | 1 |
| **iPhone** | HTTPS | HTTPS | Door unlocked with NUIT malware phished | 1000 | 1 |
| **Door** | physical | physical | Physical access to a laptop | 1000 | 1 |
| **Classified machine** | physical | physical | Physical access to state secrets | 5000 | 1 |

For establishing the parameters of the CyberBattleSim for NUIT, we build the network from the example which accompanies Microsoft's model Capture-the-Flag simulation. Table 1 shows the simulation parameters for the baseline inaudible attack network. CyberBattleSim is designed primarily for enterprise networks. Consequently, our simulation includes non-traditional nodes with services and firewall rules designated as 'physical.' These additions are essential to define the intended path through the environment, accounting for tangible actions and digital interactions. The primary reported output of the simulation uncovers a viable pathway to maximize attack rewards (e.g., a plausible attack tree). We can score the difficulty of the attack configuration by comparing the speed of conquest (e.g., all nodes owned) as measured in epoch (simulated time) steps and depth of conquest (e.g., number of nodes owned at intermediate stages). More challenging network architectures naturally require longer simulation times or remain unsolvable.

The baseline showcases the five nodes in an environment, with several states for each node. From the attackers' viewpoint, the agent is placed into a starting unknown network of properties and connectivity, followed by intermediate stages of discovery, ownership, and successful or failed actions at the nodes included in its settings for properties and node ages. The agent can execute nine possible exploits ranging from credential access to physical accesses like theft of data or equipment. Random acts at a given node may yield failed outcomes, such as physically accessing a locked smart door without the proper credential. This relatively small network offers a rich combinatorial landscape for RL (and learning algorithms) to balance exploration and exploitation.

A second set of experiments vary the algorithmic choices for reinforcement learning, as summarized in Table 2.

| Table 2. CyberBattleSim Algorithms to Balance Exploration and Exploitation in Difficult Search Environments like Networked Devices | | | |
|---|---|---|---|
| **Algorithm** | **Abbreviated Description** | **Advantages** | **Disadvantages** |
| **Random Search** | Selects actions randomly, with no regard for learned behavior | Simplicity, guaranteed to explore all possible actions | Inefficient, slow, unsuitable for large action spaces |
| **Credential Cache Lookup** | Tries to connect to random nodes using discovered cached credentials | Simplicity narrows the search space by selecting from a subset of actions | Scenarios can exist in the CBS framework that could prevent this algorithm from network ownership |
| **Tabular Q-Learning** | Off-policy RL algorithm learning Q-matrix for all state-action pairs | Effective for finite Markov Decision Processes (MDPs), simple yet powerful | Not practical for large state-action spaces ("curse of dimensionality") |
| **Exploiting Q-matrix** | Uses learned Q-matrix to guide actions, choosing the highest Q-value action in each state. | Leverages known optimal policies | No exploration might miss better solutions |
| **Deep Q-Learning (DQL)** | Uses neural network to approximate Q-function, suitable for high-dimensional state spaces | Superior performance on complex problems | Requires careful tuning, can be unstable or slow to converge |
| **Exploiting DQL** | Uses trained DQL to choose the action with the highest estimated Q-value | Improved performance on complex tasks | No exploration might miss better strategies |

## 3. RESULTS

Table 3 summarizes the end state of the CyberBattleSim, where all nodes have been discovered (stage 1) and then owned (stage 2). Using tools like CyberBattleSim can help to model, study, and devise defenses for such intricate attack scenarios, highlighting the ever-evolving complexity of the cybersecurity landscape.

| Table 3. Attack Scenario and Simulation Stages for Maximum Reward | | | |
|---|---|---|---|
| **Node id** | **State** | **Local attacks** | **Remote attacks** |
| **Echo dot** | **owned** | [Malicious Alexa Skill] | **[*NUIT2]** |
| **Email account** | **owned** | [Find Device Type In Email, Collect Data From Emails] | |
| **iPhone** | **owned** | [Unlock Door Via NUIT] | **[*NUIT1 Phishing Email]** |
| **Door** | **owned** | [Steal Classified Laptop] | |
| **Classified machine** | **owned** | [Access State Secrets] | |

This multi-stage attack underscores how various devices, some of which are not conventionally viewed as part of a network, can be chained together to enable a significant breach. Low-reward exploration is not highlighted in the final solution, which serves no real exploitation goals. For example, the smart lock on the door serves no purpose without the NUIT attack and wastes the attacker's time if mindlessly accessed from the email node. One simulation outcome underscores the exponential combinatorics of enabling a large attack surface like a typical modern home with Internet of Thing devices and VAS. As more VAS-enabled devices appear on the network map, the attacker finds a rapidly expanding attack surface just bridging between, say, Alexa to Siri to Google to Cortana. The access and enabled skills proliferate at each VAS node, including various cyber-physical assets and vulnerabilities.

The cyberattack scenario outlined in this context highlights an intricate web of interactions spanning five distinct nodes, each playing a pivotal role in the overall attack sequence. As has been demonstrated [5], the stages of the attack map to the MITRE ATT&CK matrix along with defensive remedies (MITRE D3FEND matrix).

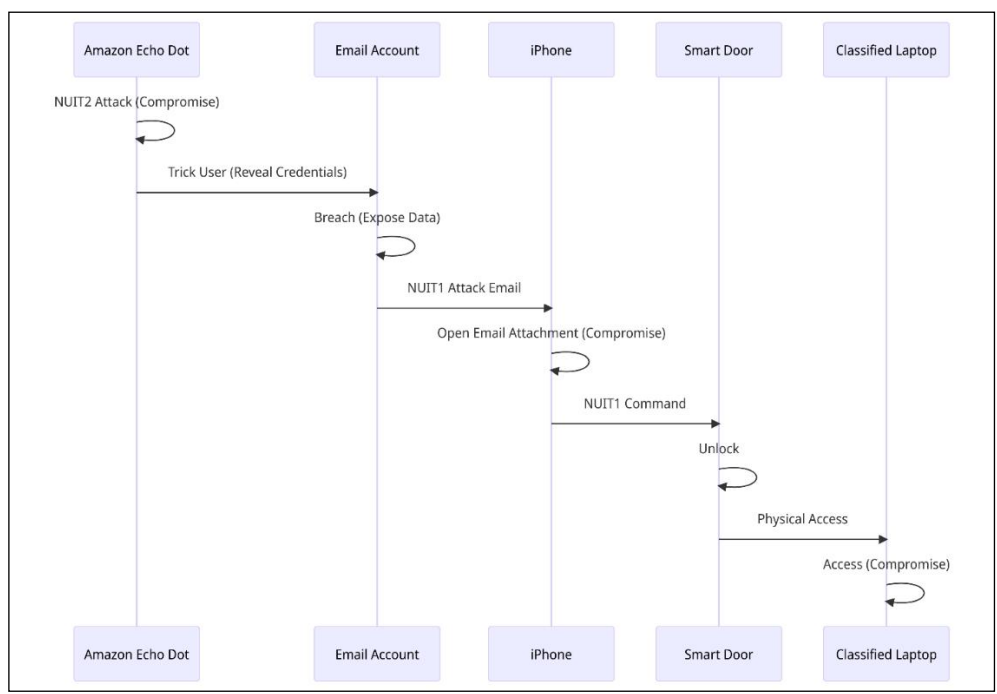

*Figure 3. Sequence diagram of the baseline NUIT Attack model under test*

- **Echo Dot:** The initial point of compromise is an Amazon Echo Dot, targeted by a NUIT2 attack that prompts it to download a malicious skill.
- **Email Account:** Once the Echo Dot is compromised, the malicious skill tricks users into revealing their email credentials. This step leads to the breach of the user's email account, which exposes sensitive personal data and provides crucial information about other potential targets within the same network, like the iPhone.
- **iPhone:** Armed with the email account's access, the attacker sends an email with a NUIT1 attack as an attachment. The iPhone, synced to the user's email, receives this email. The user unsuspectingly opens the attachment, activating the NUIT1 payload, and the iPhone is compromised.
- **Smart Door:** With control over the iPhone, the attacker commands it to send another NUIT1 command to a smart door lock, effectively unlocking the door. It's important to note that despite being a non-traditional node in the network, the smart door is a critical link in this attack chain.
- **Laptop:** The unlocked door offers the attacker physical access to the premises, where a classified laptop is located. This laptop contains valuable classified information, which now falls into the hands of the attacker.

After the initial breach of the Echo Dot, the attacker applies MITRE ATT&CK (T1202) for Indirect Command Execution. As MITRE [8] describes, "Adversaries may abuse utilities that allow command execution to bypass security restrictions…and abuse these features for defense evasion, specifically to perform arbitrary execution while subverting detections and mitigation controls (such as Group Policy) that limit/prevent the usage of command or file extensions more commonly associated with malicious payloads."

In addition to optimizing attack paths for maximum rewards in minimum time, the CyberBattleSim offers a broad survey of RL algorithms to explore in any given network, as shown in Figure 3's experimental simulations. The random search proves the worst performer even for this small network, and exploiting DQL and exploiting Q-matrix algorithms conquer the most nodes in the least time. These methods effectively work out an attack plan and execute the discovered strategy, thus maximizing exploitation as their priority and sacrificing exploration of unusual pathways or non-local maxima.

Regarding the balance between exploration and exploitation, random search excels in exploration but fails at exploitation, while epsilon-greedy offers a balance between both. Tabular Q-Learning and DQL techniques can balance both depending on the hyperparameters defined in the search strategy used during the learning phase. Still, exploitation of optimal actions maximizes rewards when using the learned Q-matrix or DQL model to make decisions.

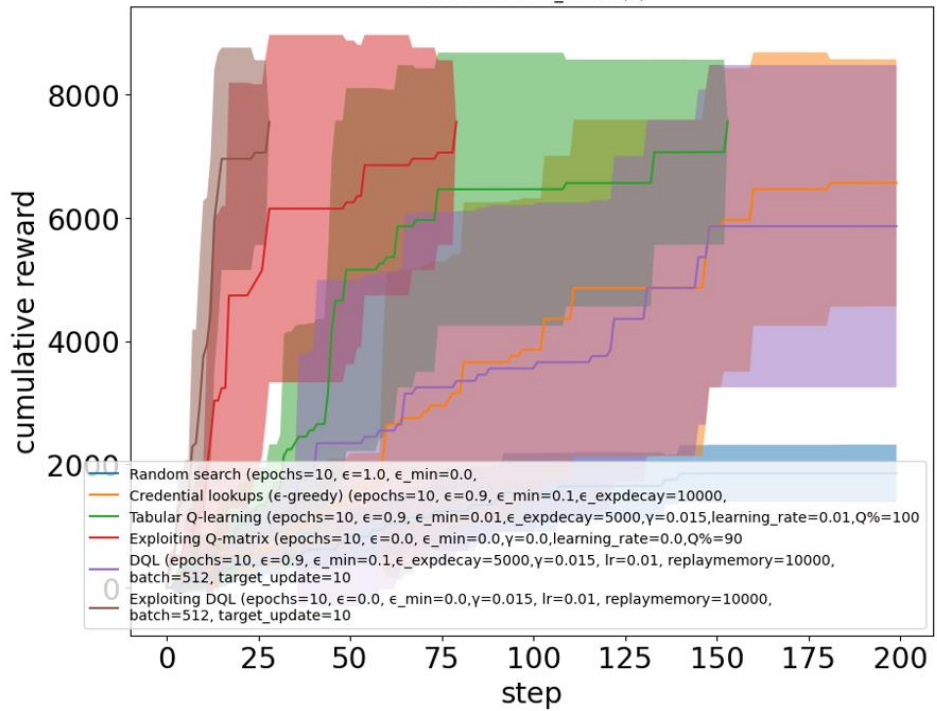

Figure 4. Effects of varying RL algorithm types and rewards accumulated over time. Better algorithms like Deep-Q Learning (DQL) accumulate rewards rapidly (upper left) compared to the worst performer as random search (blue line below rewards of 2000).

Scaling with network size and learning speed, random search does not scale well due to inefficiency. Tabular Q-Learning also struggles with scaling due to the large number of state-action pairs. DQL is most effective at scaling because it handles high-dimensional state spaces.

Regarding the risk of trapping in local minima, exploration methods (like the learning phases of Tabular Q-Learning and DQL) can potentially escape local minima. In contrast, methods focusing solely on exploitation (Exploiting Q-matrix and Exploiting DQL) risk being stuck in suboptimal solutions.

## 4. DISCUSSION

A possible shortcoming of the Cyber Battle simulation stems from its abstract network properties, which in this case, become advantageous to understanding non-conventional (disconnected) network maps and VAS attacks. A second shortcoming corresponds to its restricted coverage of lateral movements and relative inadaptability to other attack stages like reconnaissance and exfiltration. In our case, reconnaissance plays a crucial role in the simulation, which we model as a fundamental vulnerability of the VAS to reveal email and discover other assets by email credentials or signature lines.

While these simulations maintain an abstract state-action space, the CyberBattleSim on unconventional networks enables rapid coverage of possibly vulnerable network configurations in a home IoT world. The plethora of VAS configurations further motivates these tests, whether mobile phones or unrouted access points interior to another device (accessing Cortana on a laptop would not appear on a traditional network map or simulation tool like ns3 or Exata.)

In the spirit of generalizing the attack surface, we have also deferred many interesting details of how an attacker might produce an inaudible command on a VAS.

However, for specificity, we summarize the key points as requiring no special external hardware other than a voice recognition suite (audible) and an exploitable non-linear microphone for demodulating a higher frequency signal into what gets reinterpreted as an actionable attack. As first reported [2], the attack generates near-ultrasonic audio from audible sources using Single Upper Sideband Amplitude Modulation (SUSBAM).

As illustrated in time and frequency space (Figure 3), the method transforms a spoken command to a voice-activated device using the modulated audio signal converted into a frequency range (16-22 kHz) beyond human-adult hearing. Stepwise, the method applies a low-pass filter with a cutoff of 6 kHz to the original spoken signal and removes components not essential to the SUSBAM generation. The generator normalizes the remaining audio signal and applies a carrier frequency (16 kHz) to shift the audible range of the input signal to a higher frequency (near ultrasonic, 16-22 kHz). Figure 4 shows the original and transformed signal as an example in waveform (time domain) and spectral (frequency and power density domain). Figure 3 displays the audible and inaudible signals as spectrograms and how the signal gets archived after voice recognition.

In summary, the NUIT process reads an input audio signal, applies a low-pass filter, modulates the signal using SUSBAM to shift it to a near-ultrasonic frequency range, applies a Tukey window, normalizes, converts the signal to an integer format, and writes the processed signal to an output WAV file.

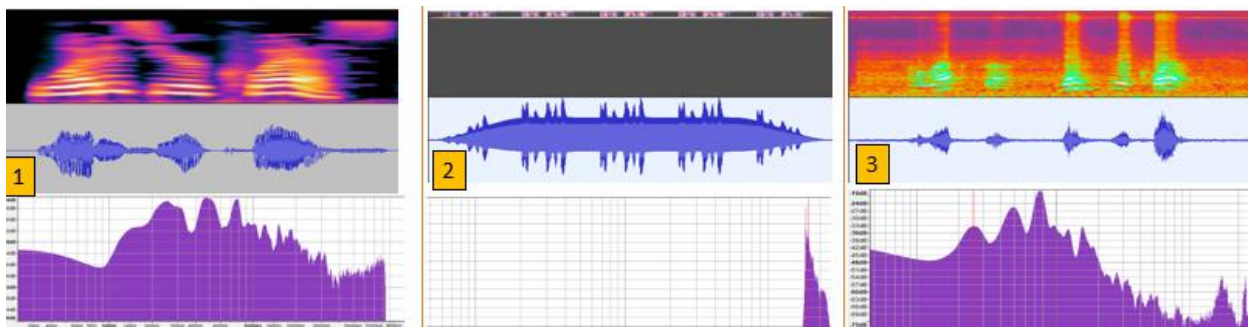

Figure 5. How the server (3) interprets inaudible broadcasts (2) transformed using NUIT from the original audible signal (1)

# 5. CONCLUSIONS AND FUTURE WORK

In this research, we addressed the significant and increasingly complex attack surface presented by voice-activated devices, which are susceptible to inaudible attacks. With a network ecosystem replete with dynamic configurations, temporal and spatial outliers, and transient targets, these devices pose intricate security challenges. We quantitatively assessed the effectiveness of stealth commands, noting a high success rate for wake word recognition and a substantial yet lesser rate for the execution of commands and controls.

This form of attack, characterized predominantly as hardware-oriented, has the potential to inflict severe damage on remotely fielded devices. It is immune to conventional defenses such as software patches or version upgrades, escalating the threat severity to 7.6 on a 10-point scale.

We proposed and implemented reinforcement learning environments like CyberBattleSim for the first time to navigate the rapidly evolving network configurations. Such environments offer a pragmatic approach to comprehending the combinatorial explosion of device types and software skills, exploring new nodes, evaluating risks, and devising mitigations.

In our study, we modeled an inaudible attack that retrieves secret information from a secured laptop, illustrating the high-stakes risks these attack vectors pose. Furthermore, we benchmarked Deep-Q learning with exploitation as a promising method for rapid learning, effective discovery, and ownership of nodes in minimal time steps. This research underscores the urgency to intensify efforts to secure our interconnected, voice-activated future.

As we advance into a future where IoT devices are increasingly ingrained into our daily lives, we must explore the intersection of these devices with voice-activated systems. A future simulation could model a heterogeneous IoT environment, encompassing smart home systems, wearable devices, and more, and scrutinize the potential of inaudible attacks across this diverse landscape.

Such a simulation could expose new vulnerabilities and attack vectors uniquely posed by the heterogeneity and complexity of IoT environments. It could also facilitate the development of novel defense strategies tailored specifically to protect these environments.

Shortcomings in this area may stem from the highly diverse and complex nature of IoT environments, potentially leading to oversights and inaccuracies in the simulation. The challenge will be creating a realistic model that accurately mirrors the vast range of device types, protocols, and vulnerabilities in real-world IoT networks.

The current model focuses primarily on the attack chain's delivery, exploitation, and action stages. Future work could delve deeper into the initial stages of an attack – specifically, reconnaissance – by modeling how attackers gather information and identify vulnerabilities before launching an inaudible attack.

These RL simulations can provide insight into how attackers plan and strategize their attacks and reveal patterns that can be used to detect and mitigate threats at an earlier stage. It will also provide a more holistic view of the entire attack process.

However, accurately modeling the reconnaissance stage can be challenging due to the many methods attackers use to gather information and because these activities often occur in difficult-to-detect and measure ways.

In the wake of a successful inaudible attack, exploring the subsequent stages of data exfiltration and persistence within the system is vital. A future simulation could model these stages to understand how an attacker can leverage a voice-activated device's control, persistently exploit it, or exfiltrate sensitive data.

Simulating these later stages of an attack can aid in understanding attacker behavior post-exploitation, potentially highlighting signs of persistent access or data breaches.

One shortcoming in modeling these stages could be the diverse nature of exfiltration methods and persistence techniques, which are often highly customized to the specific system and goals of the attacker. Additionally, these attack stages are often stealthy and difficult to detect, presenting further challenges in accurately representing them in a simulation.

In conclusion, while CyberBattleSim has provided a robust platform for investigating inaudible attacks on voice-activated devices, the future simulations described above can extend this research further. This approach will not only allow us to understand the full gamut of threats posed by such attacks but also aid in creating comprehensive countermeasures to safeguard against them.

## ACKNOWLEDGMENTS

The author would like to thank the PeopleTec Technical Fellows program for its encouragement and project assistance.